\relax
\documentclass[letterpaper]{article}
\usepackage[table]{xcolor}
\usepackage{acl2017}
\usepackage{times}
\usepackage{helvet}
\usepackage{courier}

\usepackage{latexsym}
\usepackage{color}
\usepackage{amsmath}

\usepackage{graphicx}
\usepackage{verbatim} 
\usepackage{multirow}
\usepackage{array}
\usepackage{comment}
\usepackage{tabularx}
\usepackage[font={small,it}]{caption}
\usepackage[nopar]{lipsum}
\graphicspath{ {images/} }

\aclfinalcopy
 
\begin{document}
%
\newcommand{\refalg}[1]{Algorithm~\ref{#1}}
\newcommand{\refeqn}[1]{Equation~\ref{#1}}
\newcommand{\reffig}[1]{Figure~\ref{#1}}
\newcommand{\reftbl}[1]{Table~\ref{#1}}
\newcommand{\refsec}[1]{Section~\ref{#1}}
\newcommand{\method}[1]{\mbox{\textsc{#1}}}

\newcommand{\reminder}[1]{\textcolor{red}{[[  #1 ]]}\typeout{#1}}

\newtheorem{theorem}{Theorem}[section]
\newtheorem{claim}[theorem]{Claim}

\newcommand{\system}{CANDiS}

\title{\system{}: Coupled \& Attention-Driven Neural Distant Supervision}
\author{Tushar Nagarajan \thanks{ This research was conducted during the author's internship at Indian Institute of Science, Bangalore, India.} \\
The University of Texas at Austin\\
\\
{\tt tushar.nagarajan@gmail.com}\\
\\\And
  Sharmistha and Partha Talukdar \\
  Indian Institute of Science\\
Bangalore, India\\
{\tt sharmistha.jat@gmail.com}\\
 {\tt ppt@cds.iisc.ac.in} \\
}

\maketitle

\begin{abstract}
Distant Supervision for Relation Extraction uses heuristically aligned text data with an existing knowledge base as training data. The unsupervised nature of this technique allows it to scale to web-scale relation extraction tasks, at the expense of noise in the training data. Previous work has explored relationships among instances of the same entity-pair to reduce this noise, but relationships among instances across entity-pairs have not been fully exploited. We explore the use of inter-instance couplings based on verb-phrase and entity type similarities. We propose a novel technique, \system{}, which casts distant supervision using inter-instance coupling into an end-to-end neural network model. \system{} incorporates an attention module at the instance-level to model the multi-instance nature of this problem. \system{} outperforms existing state-of-the-art techniques on a standard benchmark dataset. 

\end{abstract}

\label{sec:model}
\setlength{\textfloatsep}{5pt}
\begin{figure*}[t]
\includegraphics[width=\textwidth,height=20em]{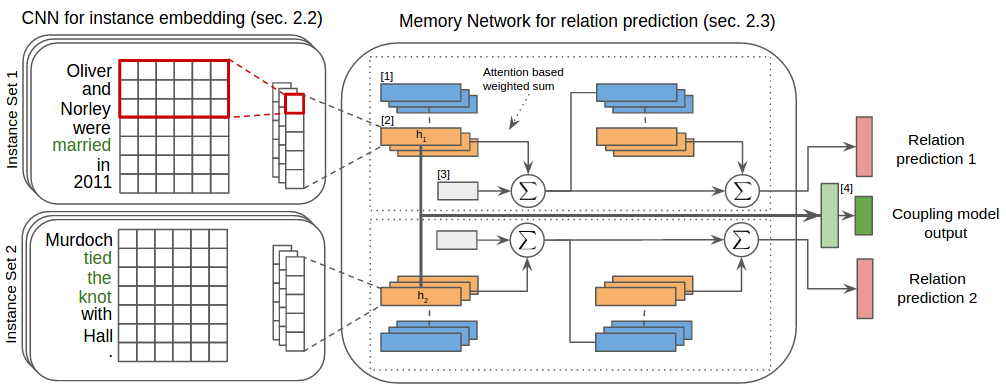}
\centering
\caption{The architecture of \system{}. Instances are embedded by CNNs into fixed length vectors. These representations are used as \textit{memory}[1] and \textit{output}[2] vectors in the standard memory network setup. The memory network uses an attention mechanism to select instances and outputs a relation label for the whole instance-set. Along with relation prediction, a coupling module[4] operates on an inter-instance level across instance-sets  which contributes another error signal in the model.}
\label{full_model}
\end{figure*}
\section{Introduction}

Relation extraction (RE) is a task in which relations expressed between entities in a sentence are discovered. Due to the scarcity of annotated data, supervised approaches to RE are not practical in a web-scale context, where free text is abundant. To tackle this problem, distant supervision is commonly employed which aligns unannotated text to a database of fact tuples, in order to generate a large volume of training data. For example, if the database contains relation `Delhi' - `Located-In' - `India', all the sentences containing entities `Delhi' and `India' will be labelled to be true for the relation `Located-In', which may not be true for some sentences. This training data is noisy, and a large proportion of the aligned sentences do not express any relation. This converts the problem of a simple classification into a multiple instance problem, and much of the previous work in RE has operated in this framework \cite{surdeanu:2012}. 

We define an instance as a sentence containing a given entity-pair, and an instance-set as set of all sentences containing the entity-pair. This instance set is the input to the model, output of the model is the set of true relations for the entity pair. We propose a model that addresses training noise on two levels. \textit{On the instance-set level}, we use memory networks \cite{sukhbaatar:2015} as an attention model to select relevant instances amidst the noise. \textit{On the instance level}, we explore various forms of couplings that allow the inclusion of global information into the representations we learn for instances. For example, sentences with semantically similar verb phrases imply same relations, this coupling is induced by training the model with multi-task objective of similarity between sentences. 

Distant Supervision (DS) for Relation Extraction was introduced by \cite{mintz:2009} using a Freebase-aligned Wikipedia corpus. A large proportion of the subsequent work in this field has aimed to relax the strong assumptions that the original DS model made \cite{riedel:2010,hoffmann:2011,ritter:2013,surdeanu:2012}. \cite{zeng:2015} proposed a Piecewise Convolutional Neural Network (PCNN) to address the issue of hand-crafted feature engineering.
In summary our contributions are:
\begin{itemize}
	\setlength\itemsep{-0.5em}
	\item We use different coupling factors on a sentence level, in a neural network based multitask framework to improve relation extraction.
	\item We use memory network proposed by \cite{sukhbaatar:2015} to reduce noise in an instance-set. 
\end{itemize}

\begin{table*}[t]

	\scriptsize
	\centering	    
		\begin{tabular}{|l|l|l|l|l|}
		\hline
		{\bf Entity-Pair}&{\bf Sentence} & {\bf Hop 1} & {\bf Hop 2} & {\bf Hop 3} \\\hline 
			\multirow{ 4}{*}{Chad Hurley - Google}&... youtube 's chief executive chad hurley received shares of google and ...&0&0&0 \\
			&... , said chad hurley , chief executive and co-founder of youtube , a division of google .&0&\cellcolor{blue!30}0.953&\cellcolor{blue!55}1.0 \\
			&google's sergey brin and larry page, skype 's janus friis , chad hurley from youtube , ...&0.0&\cellcolor{blue!05}0.041&0.0 \\
			&... , chad hurley , a youtube co-founder , ... that his site , now owned by google , ...&\cellcolor{blue!55}1.0&\cellcolor{blue!05}0.006&0.0 \\\hline
\multirow{ 4}{*}{Canada - Ontario}&.., a company in hamilton , ontario , canada , sells ...	&\cellcolor{blue!55}1.0&\cellcolor{blue!05}0.006&0.0 \\
			&she was born in ontario , canada and also lived in brazil ...&0&\cellcolor{blue!30}0.993&\cellcolor{blue!55}1.0 \\
			&canada shaw festival niagara-on-the-lake , ontario , through oct. 28 .&0.0&0.0&0.0 \\\hline
	\end{tabular}
	\label{tab:att}
	\caption{Output of the attention probabilities over an instance-set for the relations PersonInCompany(Chad Hurley, Google) and LocationInLocation(Ontario, Canada). In the first example, the model successfully selects instances 2 and 4, where there is direct evidence of the relation, while disregarding instances 3 and 4.}
	\end{table*}

\label{sec:model}

\section{\system{} for Relation Extraction}
\textbf{Approach}: Relation extraction algorithms in the distant supervision setup, take an input entity pair with set of sentences containing both the entities. The output of the model is a set of true relations between the input entity pair. Previous approaches to distant supervision treat each instance-set independent of one other during relation extraction. The assumption that each training example is completely independent however, is not strictly true. We propose a joint attention-based neural network model for relation extraction which we call Coupled \& Attention-Driven Neural Distant Supervision (\system{}). It consists of a memory network for attentive instance selection, and a coupling module to incorporate the coupling information described in \refsec{sec:coupling_specs}. We therefore cast relation extraction as a multi-task problem and leverage inter-instance coupling information to learn rich representations for instances. 

Each component of the model is described further. Complete model can be 
seen in \reffig{full_model}.

\textbf{Text Embedding} \label{sec:cnn_specs}: The instance representations are generated using a CNN following \cite{kim:2014}. The features used for each instance are word, POS-tag and position embeddings. The position features as in \cite{zeng:2014}, are integers that represent the relative distance from each entity. Two such distance vectors are produced, one for each entity involved in the relation. Each instance of $n$ tokens is therefore represented as a $ n \times (d_{word} + d_{pos-tag}+ 2\times d_{position})$ matrix, which is fed into the CNN. The  generated instance embeddings are used by memory network (\refsec{sec:memnet_specs}).

\textbf{Memory Network} \label{sec:memnet_specs}: In the distant supervision framework, many instances are labeled with a relation which is not expressed by them, leading to noise in the training data. We treat this as an instance selection problem. A memory network that iteratively selects relevant instances using an attention mechanism is ideally suited for this task. 
The end-to-end memory network from \cite{sukhbaatar:2015} is adapted for this instance selection task. The network performs $K$ passes over the instance set, and focuses on one instance in each pass. Information from these instances is then aggregated and used to predict the relation label. 

\textbf{Input Initialization} \label{sec:init_specs}: In the first iteration, naively using the zero vector as an input does not help the attention mechanism. We therefore heuristically pick simple, representative instances for each of the $|\mathcal{R}|$ relations from the training set. This is done by finding the shortest training instance that contains tokens from the relation phrase.

For example, for relation `\textit{/business/\textbf{person}/company}' we find the shortest instance with overlapping tokens is ``The latest \textbf{person} to seek assistance is the chief of [delta air lines] , [gerald grinstein] ."

We compute similarity of each candidate instance with the $\mathcal{R}$ sentences. The score of the most similar representative sentence is then used as the attention probability. This serves as an informed starting point for future iterations.

\textbf{Coupling Layer} \label{sec:coupling_specs}:
We explored several forms of coupling in this work. The most successful couplings we discovered were verb-phrase and entity-pair similarity.
\begin{itemize}
\item \textit{Verb-phrase coupling}: The verb phrases "married" and "tied the knot" are semantically related, and though they may occur in instances from different instance-sets, they represent the same semantic relation $MarriedTo(x,y)$. This consistency should be reflected in the instance representations. 

\item \textit{Entity-pair coupling}: Entity pair similarity serves as a proxy for matching entity types. Instances whose entity templates match, are likely expressing similar kinds of relations. Our instance representations should have this type-awareness as well.

\end{itemize}

In order to incorporate this coupling information into our instance representations, we clone the memory network and share all the parameters involved. This architecture is inspired by \cite{chopra:2005,mueller:2016} where it is used to compute a similarity metric between two inputs. 
In \reffig{full_model}, the instance representations ($h_1$ and $h_2$) are combined to form the final coupling output $g$.
\begin{gather} 
h_{sym}  =h_1 \cdot h_2\\
h_{asym}  =h_1-h_2\\
g  = \sigma(W_g [h_{sym}:h_{asym}]+b_g)
\end{gather}

$h_{sym}$ is the element wise product of $h_1$ and $h_2$, while $h_{asym}$ is the element wise difference. They are used to capture symmetric relations (like similarities) and asymmetric relations respectively.

For both verb and entity coupling, cosine similarity is calculated in a pairwise manner using the word embeddings of the verb or entity phrase, and the maximum of these values is used. 

Unlike our CNN's embedding parameters, which undergo task-specific fine-tuning, we use static, pre-trained Glove \cite{pennington2014glove} embeddings to calculate our similarities. Since Glove vectors are pre-trained on a separate, large-scale corpus, they capture global information about similarities that our model may fail to respect, thus regularising the network. 
 
In \reffig{full_model}, we can see that there are multiple error sources in the model now. One for the relation prediction from the memory network, and $M$ from the coupling between each instance across two instance-sets. These $M$ errors serve as intelligent regularizers on the representation that the CNN learns. 

\begin{figure}[h]
\includegraphics[width=23em,height=12em]{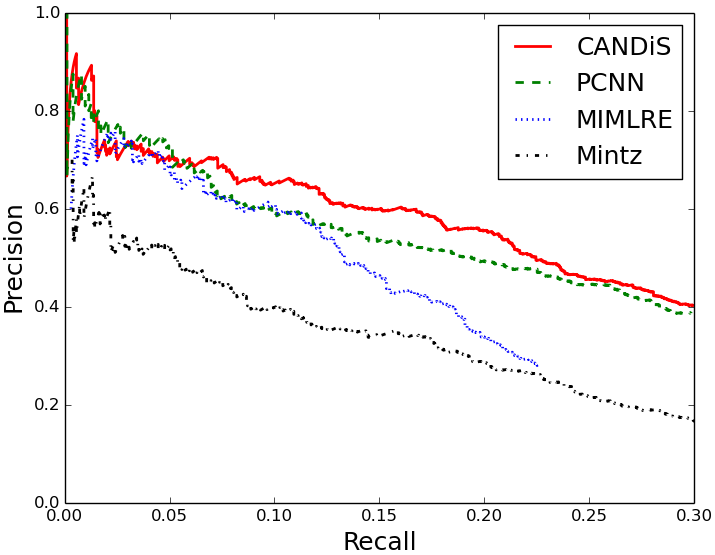}
\caption{Precision-recall curves of our model (\system{}) against traditional methods.}
\label{vs_all}
\end{figure}

\section{Experiments}
\label{sec:expts}

We evaluate our model on the dataset developed by \cite{riedel:2010} which was created by aligning Freebase relations with the NYT corpus. Our evaluation scheme is identical to \cite{riedel:2010}. Predictions are generated on a instance level and then aggregated for each entity-pair. The precision and recall at each iteration is plotted.

We follow the same protocol as \cite{mintz:2009} and evaluate our method using \textit{held-out methods,} (using distant supervision generated testing data).

\textbf{Model Parameters}:
Following \cite{kim:2014}, we use two filters (of width 1,2) in a single convolution layer, followed by max-pooling. We use $d_{word}=300$ and $d_{pos-tag}, d_{position} = 50$ for embedding our features and initialize the word embeddings using Glove vectors \cite{pennington2014glove}.
Memory network is trained over $K=4$ hops with a memory capacity of 10, and latent dimension size 256. Optimization is done using Adam \cite{kingma2014adam} with learning rate $2e-4$.

\textbf{Results}:
To evaluate our method, we compare against several competitive methods. The original distant supervision model proposed in \cite{mintz:2009}, MIML-RE proposed by \cite{surdeanu:2012} and the Piecewise-CNN model from \cite{zeng:2015}. The precision-recall curves for the held-out evaluation can be seen in \reffig{vs_all}.

\subsection{Instance Subset Selection}
The memory network architecture allows us to probe into the instance-selection process of the model. We can gauge which instances in the instance-set receive more attention from the model by visualizing the attention weight distribution. Table 1 shows these observations for a few examples. The attention mechanism manages to select only the instances that have direct evidence of a relation. Instances that exhibit very indirect or no support for the relation are often ignored. This is exactly the selection mechanism required to filter through the levels of noise in distant supervision data.

\section{Conclusion}

In this paper, we present \system{} for robust distant supervision. As we have shown, incorporating inter-instance coupling information into the the representations significantly boosts performance over a broad recall range in the relation extraction task. It would be infomative to see how this sort of coupling affects representation learning for other tasks.
As future work we would like to explore more sophisticated forms of couplings, and richer embedding models.

\bibliography{ref}
\bibliographystyle{acl_natbib}

\end{document}